# OASIS: A Deep Learning Framework for Universal Spectroscopic Analysis Driven by Novel Loss Functions

Chris Young[1], Juejing Liu[1,2,*], Marie L. Mortensen[1], Yifu Feng[1], Elizabeth Li[2], Zheming Wang[1], Xiaofeng Guo[2], Kevin M. Rosso[1], and Xin Zhang[1,*]

1. Physical & Computational Science Directorate, Pacific Northwest National Laboratory, Richland, Washington 99354, United States

2. Department of Chemistry, Washington State University, Pullman, Washington 99164, United States

**\*Corresponding Authors:** xin.zhang@pnnl.gov (X.Z.) and juejing.liu@pnnl.gov (J.L.)

## Abstract

The proliferation of spectroscopic data across various scientific and engineering fields necessitates automated processing. We introduce OASIS (Omni-purpose Analysis of Spectra via Intelligent Systems), a machine learning (ML) framework for technique-independent, automated spectral analysis, encompassing denoising, baseline correction, and comprehensive peak parameter (location, intensity, FWHM) retrieval without human intervention. OASIS achieves its versatility through models trained on a strategically designed synthetic dataset incorporating features from numerous spectroscopy techniques. Critically, the development of innovative, task-specific loss functions—such as the vicinity peak response (ViPeR) for peak localization—enabled the creation of compact yet highly accurate models from this dataset, validated with experimental data from Raman, UV-vis, and fluorescence spectroscopy. OASIS demonstrates significant potential for applications including in situ experiments, high-throughput optimization, and online monitoring. This study underscores the optimization of the loss function as a key resource-efficient strategy to develop high-performance ML models.

**Introduction**

Spectroscopy is one of the most fundamental characterization methods in fields such as science, engineering, and biology.[1-8] Recently, the generation of spectroscopic data has been exploding due to advances in spectroscopy techniques (e.g., fast and precise data acquisition),[9, 10] the demand for deeper insights and analysis of complex systems,[11] the rise of in-situ and high-throughput experiments,[12, 13] and expanding industrial monitoring applications.[14] As manually handling the overwhelming amount of data is extremely challenging, automatic data processing and analysis have become of great importance,[15] especially for in-situ/high-throughput studies,[16] operando reaction optimization,[17] and online/inline monitoring.[18] Machine learning (ML) is commonly involved in automatic data analysis as it minimizes human intervention.[19, 20] After training with suitable datasets, ML models are capable of efficiently analyzing data and obtaining results.[17] Currently, most ML-based spectroscopy processing models are tuned to solve specific problems by identifying particular features in data, for example, to distinguish chemicals and obtain crystallographic phases.[21-27]

A distinct category of ML models is aimed at the fundamental processing of spectroscopic data, including denoising, baseline correction, and peak feature extraction. While these models are not designed to answer specific analytical questions directly, they replicate the routine data processing performed by human experts—automatically and with significantly greater speed and consistency. Previous studies have developed automated processing solutions tailored to specific techniques such as Raman and nuclear magnetic resonance (NMR) spectroscopy. In Raman applications, ML has primarily focused on denoising raw spectra, with the cleaned outputs serving as inputs for downstream analyses.[28-30] For NMR data, where noise is typically less severe, ML models have been used to identify peak positions and extract parameters like intensity and full width at half maximum (FWHM), even in complex or overlapping peaks.[31-33]

However, a general-purpose, full-stack pipeline that performs all preprocessing and peak extraction tasks across various spectral types has not yet been realized. Our work addresses this gap by pursuing two core objectives: (1) developing technique-independent models for broad applicability across spectroscopies, and (2) creating computationally efficient architectures that can be trained on relatively small datasets. These models are designed to operate on standard laboratory computers and potentially edge devices, eliminating the need for specialized ML hardware such as GPUs. By minimizing both training data requirements and model complexity, we aim to deliver practical tools for real-world spectroscopic workflows.[34, 35, 36-39]

In this study, we present OASIS (Omni-purpose Analysis of Spectra via Intelligent Systems), a U-Net-based, technique-independent framework for fully automated spectroscopy data processing. OASIS consists of four specialized models for denoising, baseline correction, peak location detection, and retrieval of peak intensity and FWHM. Technique independence is achieved through training on a strategically designed synthetic dataset that mimics features across diverse spectroscopy techniques. To enhance efficiency, we introduce task-specific loss functions, including total variation loss for baseline removal, ViPeR loss for peak localization, and mean cubic/cubic error (MCE/MQE) for peak intensity and width prediction. These strategies enable compact yet powerful models, with the largest having only 438,422 trainable parameters. OASIS demonstrates strong extrapolation capabilities, accurately processing features not seen during training—such as removing spark-induced noise and detecting more peaks than included in the training set—making it a robust and versatile tool for real-world spectral analysis.

**Results**

As illustrated in **Figure 1a**, OASIS uses a series of U-Net-based models to sequentially process spectra, regardless of the original spectroscopy technique. These processes encompass: (i) denoising, (ii) baseline correction, (iii) peak location identification, and (iv) peak intensity and FWHM retrieval, where each subsequent model in the pipeline utilizes the output from the preceding stage. To train these models, we strategically designed a synthetic spectral dataset that incorporates a wide array of features representative of various spectroscopy techniques (see **Methods** section). For each model (**Figure 1b**), three distinct data streams are fed into three parallel U-Net-based encoder-decoders, the spectrum itself, along with its first and second derivatives. The outputs from these three U-Nets are then concatenated to produce the final prediction (see **Methods** section). This multi-input approach provides a comprehensive representation of the spectral data to the model, thereby enhancing the accuracy of the various processing stages. Furthermore, we developed a set of customized loss functions dedicated to different tasks, for example, ViPeR loss (**Figure 1c**) to train the models for effective peak localization. Compared to the commonly used cross-entropy loss,[31, 32] our ViPeR loss offers a smoother loss landscape, which facilitates more efficient training for identifying peak locations. This characteristic is intended to guide the optimization towards more robust and generalizable minima than those potentially encountered with conventional cross entropy (CE) loss.

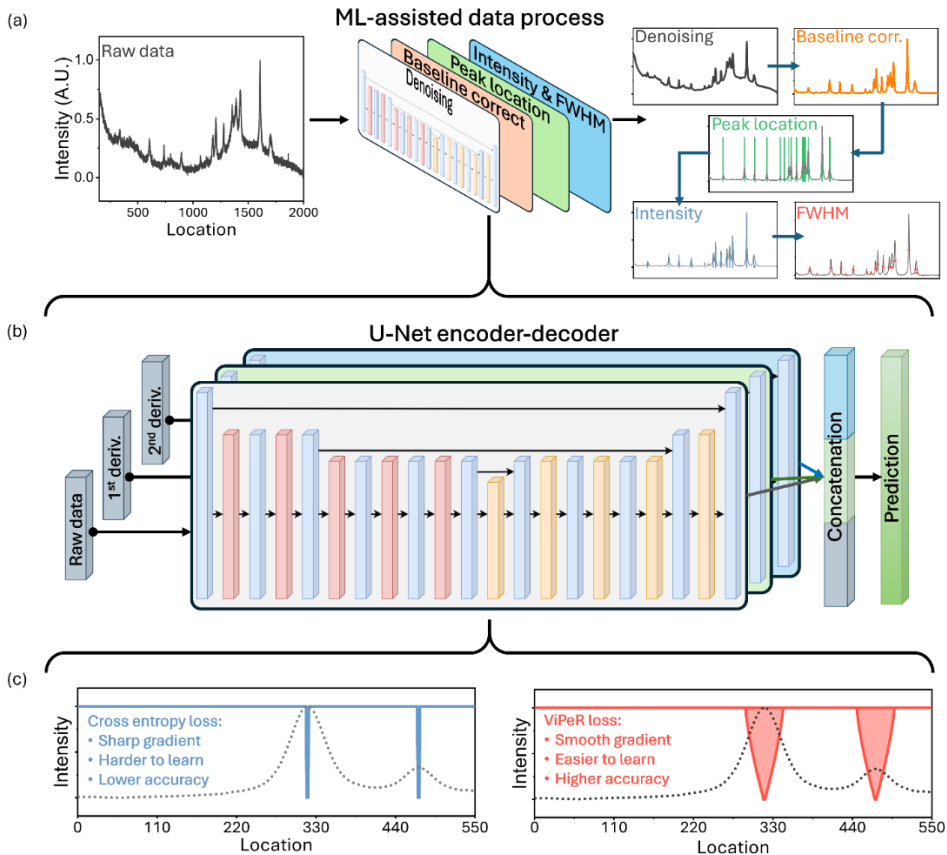

**Figure 1**. Illustration of the OASIS workflow capable of processing spectroscopy data regardless of the type of technique. (a) The four models comprising OASIS, including denoising, baseline correction, peak location detection, peak intensity and FWHM retrieval. The raw data are processed by the above model sequence to obtain useful information. (b) The U-Net architecture used in all four models. (c) Comparison of conventional cross-enthalpy loss and vicinity peak response loss.

**Peak Identification with Novel Loss Function**

To improve training efficiency and accuracy for spectral peak identification, we introduce ViPeR loss. Peak identification is typically framed as a point-wise binary classification problem using CE loss,[31,32] but CE often converges toward sharp minima that impair generalization and robustness.[40,41] A key challenge is severe class imbalance, where non-peak points vastly outnumber peak points, hindering effective optimization.[42] While data augmentation may help, it can introduce unrealistic spectral features.[43] Standard approaches combine CE for classification and MSE for parameter prediction,[29,31,37,44,45] but CE performs poorly on imbalanced datasets.[42,43] The ViPeR loss provides smooth gradient signals near spectral peaks, enabling networks to first learn general peak patterns then refine localization accuracy. Models trained with ViPeR loss consistently outperformed those using conventional CE+MSE loss across evaluation criteria, with accuracy improvements from 0.889 to 0.932 under optimal conditions (**Figure 2**) (For further details see **Methods**).

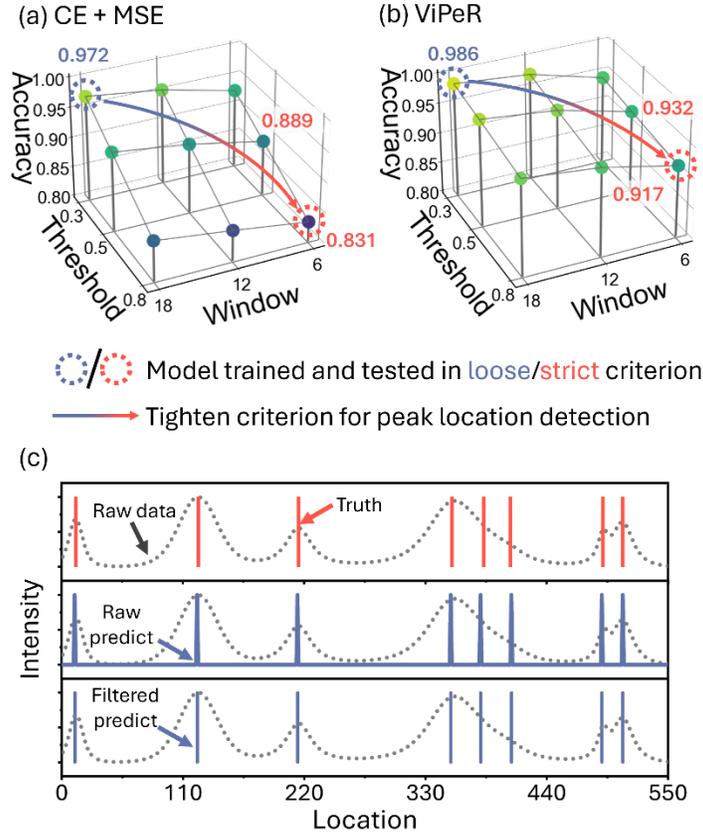

**Figure 2.** Comparison of the traditional CE and MSE loss function and the ViPeR loss function in identifying peak locations. (a) and (b) Change of test MSE vs. the two criteria that successfully identify peaks, including probability threshold and window size for the cross-entropy loss function (a) and the ViPeR loss function (b). (c) Demonstration of the output of the peak location identification model trained by the ViPeR loss function.

To evaluate the impact of the loss function on model performance, two U-Net models were trained: one using the ViPeR loss and another using a conventional combination of CE and MSE loss. Preliminary trials indicated that providing the model with the raw spectral data along with its first and second derivatives yielded optimal training performance, evidenced by convergence to a lower loss value (see **Method** session and **Figure 1c**). The accuracy of peak location identification was assessed based on two criteria: the confidence threshold for peak existence (derived from the sigmoid output) and the window size used in post-processing non-maximum suppression (NMS). Accuracy is defined as the ratio of correctly predicted peak locations to the total number of ground truth peaks. This definition focuses on true positives and false negatives, as the peak location models developed in this study were observed to exhibit a low propensity for false positive predictions.

**Figures 2a**, **2b,** and **Table S2** demonstrate that the ViPeR loss function yields superior performance compared to the conventional use of CE with MSE, irrespective of the post-processing criteria. Under loose criteria (e.g., a confidence threshold as low as 0.3 and an NMS window size as large as 18 slots), the model trained with ViPeR loss (0.986 precision) performed noticeably better than the model trained with CE and MSE loss (0.972 precision). It is important to note, however, that such loose criteria, despite yielding an apparently higher metric, may impair model performance on experimental data by potentially causing false positive peak detections or failing to resolve convoluted peaks due to the large NMS window. As the criteria became stricter (higher threshold and smaller NMS window), the performance gap between the ViPeR loss-based and CE with MSE-based models widened. For example, with a threshold of 0.5 and an NMS window of 6, the maximum identification accuracy of the ViPeR loss-based model was 0.932, while the accuracy for CE with the MSE loss-based model decreased to 0.889. Under the strictest criteria tested (threshold of 0.8 and NMS window of 6), the precision of the ViPeR loss-based model was maintained at 0.917, while that of the CE with MSE loss-based model further decreased to 0.831. These results collectively indicate that models trained using ViPeR loss exhibit consistently better accuracy across a wide range of evaluation criteria compared to those trained with a CE and MSE loss combination. Taking into account the balance between optimal peak identification accuracy and the ability to handle convoluted peaks, a confidence threshold of 0.5 and an NMS window size of 6 were adopted for subsequent studies. The resulting model is capable of identifying isolated and convoluted peaks from spectral data, as demonstrated in **Figure 2c** using a synthetic spectrum.

**Denoising and Baseline Correction**

In the majority of spectroscopic preprocessing workflows, the initial steps typically involve denoising and baseline correction. Consequently, the automation of these two processes is crucial before spectral data is passed to subsequent models, such as those for peak location detection. Although previous studies have proposed ML-based denoising and baseline correction models for specific types of spectroscopy such as Raman, UV-Vis, and fluorescence spectroscopy,[38, 46-49] our work, inspired by these earlier efforts, focused on enhancing the versatility of our denoising and baseline correction models to accommodate a wide range of noise levels and baseline shapes (**Figure 3**).

To enhance the versatility of the model to handle varying noise levels, a weighted MSE loss function was used during training (**Figure 3a**).[50-53] The synthetic spectral datasets used for training the noise removal model incorporated five distinct noise levels, with standard deviations ranging from 0.001 (minimal) to 0.15 (most severe, see details in the Method section). During training, the MSE loss, calculated between the ground truth and the prediction of the model, was modulated by a scaling factor. For spectra with severe noise, this scaling factor was set to 1, while for spectra with minimal noise, the MSE loss was amplified by a factor as high as 100. This strategy was designed to encourage the model to learn to process a broad spectrum of noise intensities effectively. Models trained using the weighted MSE consistently exhibited lower exam MSE values in the unseen test dataset compared to models trained with a pristine (unweighted) MSE loss (**Figure 3a** and **Table S1**). The MSE exam for the model trained with pristine MSE loss was $1.32 \times 10^{-4}$. Applying even a relatively moderate weighting factor (e.g., 3) to the low-noise data improved model performance, resulting in an exam MSE of $1.23 \times 10^{-4}$. Although further increasing the weighting factor for low-noise data continued to improve the exam MSE, these improvements showed diminishing returns. Ultimately, a set of weighting factors was selected, 100, 40, 15, 4, 1, and 1 for noise levels corresponding to standard deviations of 0.001, 0.01, 0.02, 0.05, 0.1, and 0.15, respectively. The model trained with this optimized set of weighting factors achieved an exam MSE of $1.19 \times 10^{-4}$.

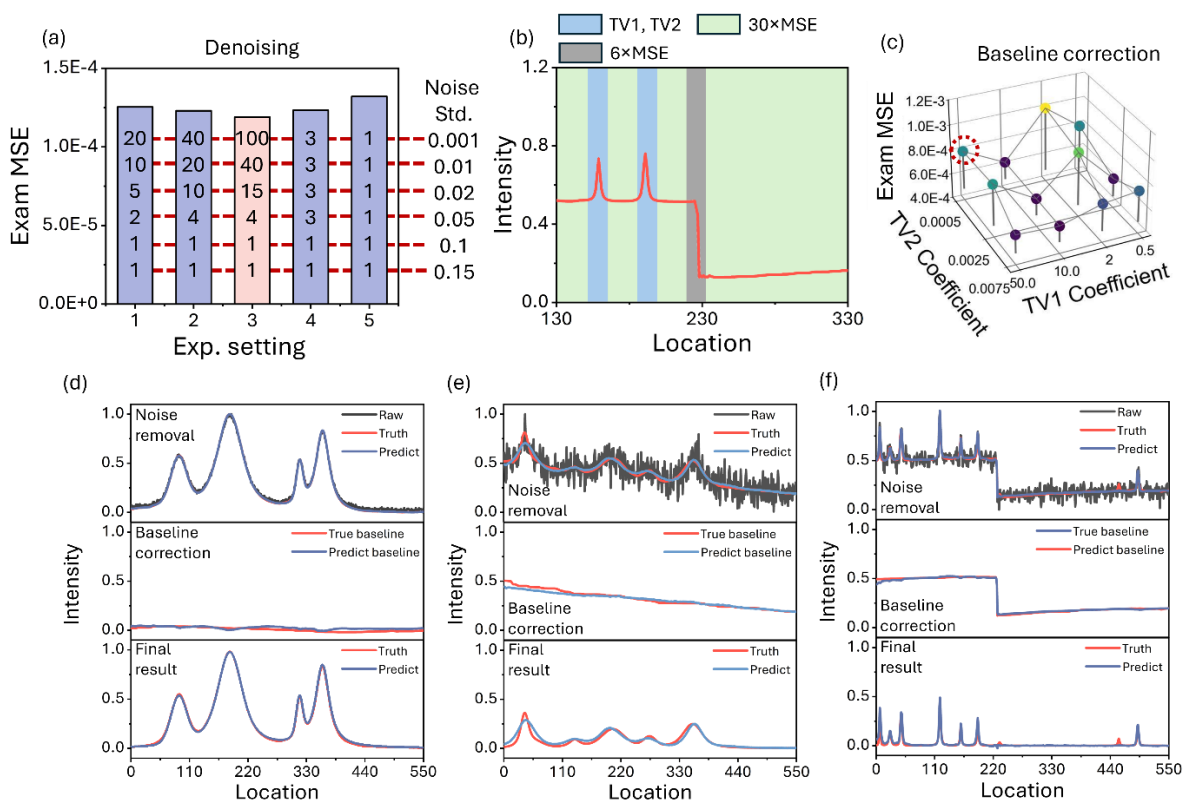

**Figure 3.** Development of U-Net-based denoising and baseline correction models. (a) Optimization of weighted MSE parameters versus noise level to lower exam MSE. (b) Illustration of the dominant contribution of loss functions in different regions of the synthetic spectrum including the loss of total variation of first order (TV1), the loss of second order variation (VT2), and MSE with different factors. (c) Influence of α in TV1 and β in TV2 on models in terms of MSE exam MSE. The red circle shows the models used in later sections of the study. (d) to (f) Demonstration of the denoising and baseline correction model in moderate noise and baseline (d), severe noise and baseline (e) and server noise with shifted baseline (f).

Spectroscopic baseline correction models often struggle with erratic baseline behavior, particularly in regions of high uncertainty such as beneath spectral peaks, leading to oscillatory and unreliable predictions. Therefore, an effective baseline correction model should be designed to: (1) encourage and maintain general baseline trends even in uncertain (e.g., peak-containing) regions, (2) aggressively remove the baseline in clearly identifiable peak-free regions, and (3) accurately accommodate and model abrupt baseline shifts. Our approach is tailored for removing artifactual backgrounds, rather than modeling the physically meaningful baselines found in certain spectroscopic techniques and substances.

Training neural networks for spectroscopic baseline correction using only MSE loss often yields erratic baseline predictions, particularly beneath spectral peaks where model uncertainty is highest. This can result in oscillations where the predicted baseline alternates between over- and under-shooting the true baseline, which distorts the recovered signal (see **Figure S4**). To address this, we developed a composite loss function incorporating three key mechanisms (see **Figure 3b**). First, a total variation loss term ($TV1 = \alpha \sum |y^{(i+1)} - y^{(i)}|$) promotes overall baseline smoothness. Although beneficial, TV1 alone is not sufficient to prevent localized fluctuations in peaks.[54-56] Second, we introduced a square root second-derivative penalty ($TV2 = \beta \sum \sqrt{|y^{(i+2)} - 2y^{(i+1)} + y^{(i)}|}$).[57] This term predominantly penalizes small, high-frequency curvature changes, effectively suppressing noise, while having a proportionally smaller impact on larger, legitimate baseline variations, thereby preserving natural concavity and avoiding the piecewise linear predictions characteristic of an unmodified second-derivative penalty. The coefficients α and β, which balance these first- and second-order regularization terms, were optimized to maintain appropriate baseline concavity. However, these smoothing penalties can impair the model's performance in two critical scenarios: aggressive baseline removal in peak-free regions and accurate modeling of abrupt baseline shifts. To counteract this, we implemented a targeted loss weighting. For spectra that resemble Raman data with well-defined peaks (indicative of clear peak-free regions elsewhere), the MSE loss contribution was amplified by a factor of 30 (30×MSE), enabling the model to selectively override smoothing constraints. Furthermore, for identified baseline shifts, smoothing penalties are suspended within a ±3 data point window around the shift, and the MSE contribution within this window is simultaneously increased six-fold (6×MSE). This approach allowed the model to accurately reproduce discontinuities without imposing inappropriate smoothing.

To optimize the behavior of the *TV1* and *TV2* components in the composite loss function for baseline correction, the models were trained with a wide range of α (from 0.5 to 50) and β (from $5 \times 10^{-4}$ to $7.5 \times 10^{-3}$) combinations (**Figure 3c** and **Table S1**). Adjusting α and β significantly impacted the MSE of the models; the difference between the MSE of the highest observed exam MSE ($1.01 \times 10^{-3}$) and the lowest ($6.21 \times 10^{-4}$) was 38%. In particular, the model that showed the absolute lowest exam MSE ($6.21 \times 10^{-4}$) demonstrated reduced capability in handling spectra with multiple convoluted peaks (**Figure S4**,). This particular model showed a higher tendency to predict oscillating baselines, where the predicted baseline would drift below or above the true baseline. It is hypothesized that an excessive contribution from TV2 in this model overmodulates small fluctuations in the prediction, which, while aiming to preserve the recovered signal, can lead to these undesirable artifacts. After careful examination of the performance of the baseline correction models against various peak characteristics, a model with an exam MSE of $8.03 \times 10^{-4}$ was selected for subsequent studies, as it offered a better balance between quantitative error and qualitative performance in challenging spectra.

The selected noise removal and baseline correction models were subsequently utilized to process three synthetic datasets simulating typical spectral features. **Figure 3d** illustrates a spectrum with a low noise level and large, broad peaks. In this case, both the noise and the baseline were successfully corrected by the respective models, yielding a final result nearly identical to the ground truth. While this represents a relatively straightforward scenario, the synthetic spectrum depicted in **Figure 3e** presents a more challenging case with a high level of noise and smaller and broader peaks. The baseline correction model accurately corrected for the baseline and the denoising model successfully reconstructed most of the peaks. A notable exception was the first peak, located at approximately position 55. Although the denoising model preserved the peak's location, its predicted shape was slightly broader than the ground truth. This outcome is anticipated when reconstructing spectra with high noise levels, particularly since the denoising model was tuned to suppress high-frequency noise, and the first peak in this example contained a significant spark-like feature that triggered this strong noise suppression. Although further tuning might address this specific instance, there is a risk of overfitting the model. **Figure 3f** presents a spectrum with relatively sharp peaks and a significant baseline change. The denoising model performed well, removing noise while preserving most of the peak structures. The two peaks at positions 232 and 456, which were exceptions, had low intensities comparable to the noise level, indicating that the denoising model is unlikely to misidentify noise as genuine peaks, a crucial attribute for processing noisy data. The baseline correction model also successfully identified and corrected the pronounced baseline shift, further demonstrating the effectiveness of the composite loss function strategy.

**Peak Intensity and FWHM Retrieval**

The intensity and FWHM of spectral peaks are key parameters in spectroscopic analysis. A major challenge in accurately determining these values arises from peak convolution, which can result in features such as shoulder peaks and broadened peaks. In ML-assisted spectrum analysis, a common approach involves building regression models. Consistent with previous methodologies,[31, 32] our model for intensity and FWHM retrieval utilizes the raw spectral data, its first and second derivatives, and the identified peak locations as inputs. Each of these input streams is processed by a dedicated U-Net block, and the resulting feature maps are subsequently concatenated. The model's output for intensity and FWHM consists of data arrays with the same number of points as the input spectrum. Predictions of intensity and FWHM are made at the data points corresponding to the peak apexes; for all other points, the model outputs a diminished value close to zero. In our preliminary trials employing a conventional MSE loss function to train the U-Net model, we observed suboptimal MSE values for the prediction of FWHM in the exam dataset, indicating the need for further refinement of the loss function.

To address this issue, we introduced the mean cubic error ($MCE = \frac{1}{n}\sum_{i=1}^{n}(y^i - \hat{y}^i)^3$) and mean quartic error ($MQE = \frac{1}{n}\sum_{i=1}^{n}(y^i - \hat{y}^i)^4$). This decision was based on the hypothesis that the suboptimal FWHM prediction performance with MSE stems from the model overfitting to the training data. When the predicted value is close to the ground truth, MSE loss can still generate a relatively large error signal, potentially forcing the model to overfit. MCE and MQE, on the contrary, are designed to provide progressively milder error signals as the prediction approaches the ground truth (**Figure S5**).

Our training strategy employed a dynamic loss function. In the initial training stages, when the difference between predicted and ground truth intensity/FWHM is large, a 5×MSE loss was applied to data points corresponding to peak apexes, while a standard MSE loss was applied to nonpeak data points. As the predictions for intensity and FWHM approached ground truth within a defined relative error threshold ($T_{MCE}$) defining by $T = \frac{|y_P^{(i)} - y_T^{(i)}|}{y_T^{(i)}}$, the 5×MSE loss at peak apexes was replaced by a 5×MCE loss, providing a gentler error gradient. Subsequently, if the predictions further converged within a stricter threshold ($T_{MQE}$), the loss term was switched to 5×MQE, offering an even milder error. This loss function configuration is formulated to incentivize the model to generate robust and balanced predictions, particularly when analyzing regions with convoluted or overlapping spectral peaks, encouraging effective estimation of underlying signals amidst complex interferences. On the contrary, in spectral regions characterized by clearly defined and well-resolved peaks, where the signal is less ambiguous, the initial dominance of MSE ensures that the loss function prioritizes predictive accuracy by more significantly penalizing deviations from the true signal.

The two thresholds, $T_{MCE}$ (ranging from 0.70 to 0.98 relative to ground truth) and $T_{MQE}$ (ranging from 0.90 to 0.99), were adjusted to optimize the precision of the intensity and FWHM predictions (**Figure 4a** and **Table S3**). This optimization process had a negligible impact on the accuracy of the peak intensity predictions; the difference between the MSE of the highest exam MSE (6.43 × 10$^{-3}$, when MCE and MQE were not applied) and the lowest exam MSE (6.41 × 10$^{-3}$ with $T_{MCE}$ = 0.94 and $T_{MQE}$ = 0.97) was merely 0.2%. This minimal change is understandable, as intensity prediction is a relatively straightforward task. On the contrary, the accuracy of the prediction of FWHM was significantly enhanced by $T_{MCE}$ and $T_{MQE}$ optimization. The MSE of the exam for the prediction of FWHM of the trained model without this dynamic loss function strategy was $2.22 \times 10^{-4}$. The optimized model, employing $T_{MCE} = 0.94$ and $T_{MQE} = 0.97$, achieved an exam MSE of $1.76 \times 10^{-4}$, representing a 21% improvement in terms of exam MSE. Consequently, the model trained with $T_{MCE} = 0.94$ and $T_{MQE} = 0.97$ was selected for all subsequent studies.

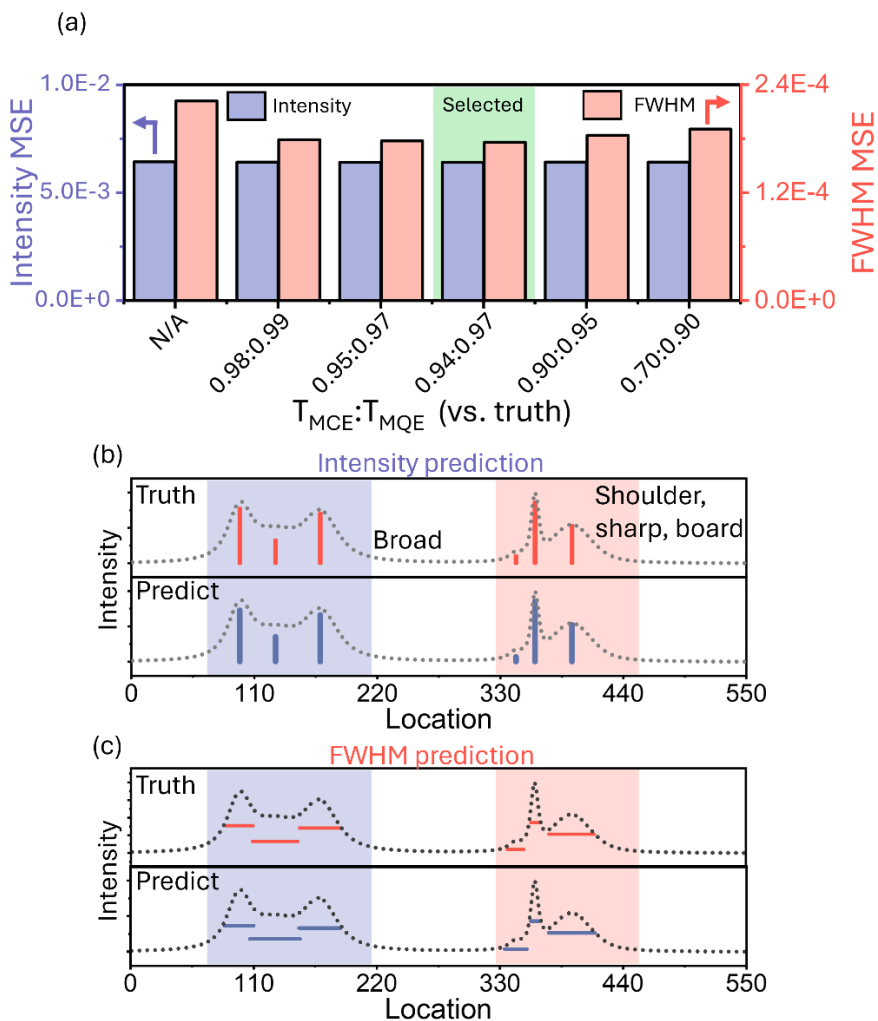

**Figure 4.** Development of peak intensity and FWHM retrieval models. (a) Optimization of two thresholds, $T_{MCE}$ and $T_{MQE}$, allowing MCE loss and MQE loss during model training. See **Table S3** for details. (b) and (c) Demonstration of the peak intensity (b) and the FWHM retrieval (c) model.

The predictive capabilities of the optimized intensity and FWHM model were demonstrated using a synthetic spectrum containing two distinct sets of typically convoluted peak characteristics (**Figures 4b** and **4c**). The first series of three peaks, located in the region from position 100 to 220, represented a convolution of broad peaks. The second series, which covered positions 330 to 440, presented a more complex convolution involving a broad, sharp, and small shoulder peak. The model successfully identified the intensity and FWHM for all peaks in both series, including the challenging small shoulder peak in the second series. These results confirm that the developed intensity and FWHM model is capable of accurately handling spectroscopic data that exhibit a wide range of peak features and convolutions.

**Experimental Validation Results**

The practical utility of the OASIS pipeline was validated on diverse experimental datasets, assessing its performance on real-world Raman, fluorescence, and UV-Vis spectra (**Figure 5**). The fully automated pipeline was applied sequentially without manual input, demonstrating its capacity for high-throughput processing. OASIS demonstrated remarkable capability across all cases. For a complex Raman spectrum (**Figure 5a**), it showed strong extrapolation, successfully identifying all 18 molecular peaks despite being trained on synthetic data with a maximum of 11 peaks. In fluorescence analysis (**Figure 5b**), the pipeline proved robust against un-trained noise, successfully removing spark-induced artifacts. This case also highlighted the resilience of the modular design: even when the baseline correction model did not fully resolve an ambiguous feature, the downstream peak detection model correctly identified it as baseline, preventing a cascading error. Finally, for UV-Vis spectra with severe artifacts (**Figure 5c**), OASIS effectively corrected significant and abrupt baseline jumps, a common and difficult challenge for automated processing, to produce clean, usable data. In all examples, quantitative outputs showed strong agreement with manual expert analysis, establishing OASIS as a reliable tool for diverse spectroscopic applications (for detailed analysis see **Methods**)

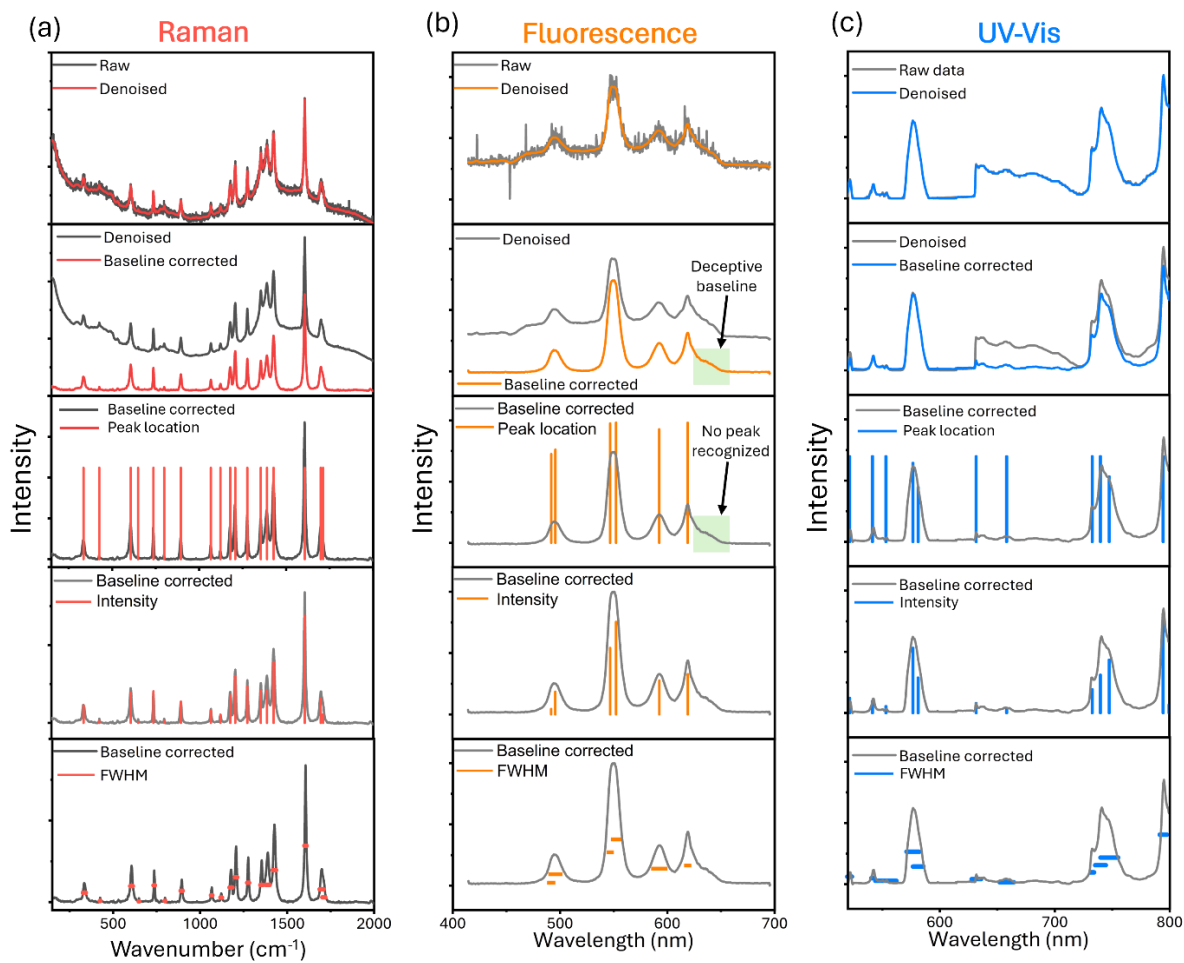

**Figure 5.** OASIS spectroscopy analysis system to denoise, correct baseline, identify peak location, retrieve peak intensity, and FWHM from experimental data from (a) Raman spectroscopy of 0.5 M 9-fluorenone-2,7-dicarboxylic acid in 3M KOH, (b) fluorescence spectroscopy of 0.04 M $TbCl_3$ aqueous solution, and (c) UV-vis spectroscopy of 0.08 M $Nd_2(SO_4)_3$ aqueous solution.

**Discussion**

  This study highlights two critical strategies for developing versatile and accurate OASIS models: building a comprehensive synthetic spectral dataset and designing customized loss functions. While the value of large datasets in ML is well established, a key contribution of this work is demonstrating how carefully engineered, task-specific loss functions significantly improve model accuracy and specificity. We showed that loss functions tailored to individual tasks—such as ViPeR for peak localization, combined TV and weighted MSE for baseline correction, and dynamic error terms (MCE/MQE) for peak parameter prediction—yielded high-fidelity results and minimized overfitting. These custom functions provided more informative feedback during backpropagation than standard MSE or CE losses, leading to superior model performance. Importantly, OASIS is designed to prevent error propagation across stages; for example, baseline artifacts (**Figure 5b**) do not impact peak prediction.

This work illustrates that strategic loss function design is a powerful and underutilized approach to enhancing ML model performance—often more effective than increasing dataset size or model complexity, with benefits including faster training and lower computational demands. Despite its overall strong performance, some limitations were observed. In cases of extremely noisy signals, the denoising model occasionally distorted broad peaks (**Figure 3e**). Nonetheless, the OASIS pipeline is computationally efficient, capable of processing typical spectra in seconds using standard laboratory computers without GPUs and is potentially deployable on edge devices. Additionally, each of the four deep learning models within OASIS can be repurposed for general spectroscopic analysis. With minimal fine-tuning, they can be adapted to specific techniques and analytical tasks. Together, these results underscore the robustness, efficiency, and broad applicability of OASIS for automated, reliable spectral analysis across diverse experimental settings.

**Conclusions**

In conclusion, this study presents OASIS, a deep learning–based tool for fully automated spectroscopy data processing, including denoising, baseline correction, and extraction of peak parameters (location, intensity, FWHM) across diverse spectral types without human input. OASIS's performance is driven by a dual strategy: training on a broad synthetic spectral dataset representing multiple spectroscopies, and the use of task-specific loss functions. Innovations such as ViPeR for peak localization and tailored combinations of weighted and dynamic losses for denoising and parameter prediction were critical to achieving robust U-Net performance. Experimental validation on Raman, UV-Vis, and fluorescence spectra showed that OASIS can deliver expert-level results within seconds. Designed for use on standard laboratory computers without specialized acceleration hardware, OASIS is versatile, accurate, and computationally efficient—making it a powerful tool for high-throughput, autonomous spectral analysis across research and industry.

**Methods**

*Overall summary*

OASIS, an automated spectral information system, is architected as a four-network pipeline sequentially performing denoising, baseline removal, peak detection, and peak fitting (intensity/FWHM), a modular design chosen after extensive experimentation demonstrated its superiority over monolithic approaches due to the highly task-specific optima identified for network configurations, hyperparameters, and particularly, loss functions (detailed in *Model Architectures and Development*). A cornerstone of OASIS's success lies in the generation of an exceptionally rich and realistic synthetic training dataset. This dataset featured a sophisticated probabilistic multi-stage baseline drift model, capable of emulating a wide gamut of complex experimental phenomena (**Figure S1**) far exceeding the capabilities of simpler literature methods, thereby exposing the networks to a comprehensive representation of real-world conditions. Crucially, this was complemented by a dual Gaussian and beta distribution-derived noise model, specifically designed to simulate not only common instrumental noise but also subtle, localized spectral artifacts or 'dents' observed experimentally, which significantly enhanced model robustness and mitigated overfitting. While all networks utilized a U-Net foundation and benefited from a multi-channel input (raw spectrum, first and second derivatives, as justified in **Figure S2**), these tailored loss functions, in concert with the rich synthetic data and targeted augmentations like spectral tweaking and peak shadowing, were critical in enabling each specialized network module to achieve high accuracy and robustness.

The second major achievement was the development and implementation of custom and dynamically weighted loss functions, which demonstrably improved performance beyond standard approaches. For instance, the denoiser's dynamically weighted MSE loss was pivotal in preserving detail in low-noise spectra (**Figure S3**); the baseline remover's composite MSE with a smoothing term ensured smoother, more realistic baseline predictions under peaks on experimental data (**Figure S4**); and a specialized loss function for the peak fitter directly addressed and corrected the tendency of standard MSE to under-predict peak intensities, leading to more accurate quantification and improved FWHM prediction (**Figure S5** and **Figure S6**). The resulting system, whose performance on well-resolved peaks shows good agreement with human annotations (**Table S4**), thus represents a significant advancement in automated spectroscopic analysis, largely attributable to these key innovations in data generation and learning optimization.

*Synthetic training data generation*

**Generation strategy**. Synthetic spectra were generated to train and evaluate OASIS, a process involving the creation of Voigt profile-based peak structures followed by the addition of a complex, probabilistically generated baseline drift (see **Figure S1**). The synthetic spectra with 555 datapoints were generated to train and test the models. Each spectrum was constructed as a sum of 1 to 11 individual Voigt profiles, with an equal representation of spectra for each count within the training dataset. For each Voigt profile, the amplitude and Full Width at Half Maximum (FWHM) were randomly generated, and all profiles were unique, with no reuse across different spectra. The center wavelength for each peak was randomly assigned; to simulate varying degrees of peak overlap, 50% of generated spectra had constituent peaks constrained to a maximum separation of 2 datapoints, while for the remaining 50%, this maximum separation was 12 datapoints.

Subsequently, a multi-stage probabilistic model was employed to simulate realistic baseline drift, informed by observations of diverse experimental spectra and designed to overcome limitations of simpler polynomial-based simulations. This is because the polynomial baseline is insufficient for robust neural network performance on baseline correction, particularly at spectral peripheries. For a given spectrum, this baseline generation began with the assignment of a global slope coefficient. The spectral wavelength range was then divided into a number of primary regions as an integer randomly sampled from 1 to 10. The initial length of each region, (Total Wavelength Range) / (Number of Primary Regions) is multiplied by a factor from 0.5 to 2 (uniformly selected) introducing a slight bias towards larger segments while permitting erratic variations; baselines were truncated or extended if the cumulative adjusted lengths overflowed or underfilled the spectral range. Each primary region had a 15% probability of being flat, overriding subsequent slope calculations. If not flat, a region had a 50% chance of an overall positive or negative slope, and a regional slope multiplier from 0.01 to 1.99 was uniformly selected to scale the global slope coefficient (multiplication). Each primary region was further subdivided into smaller 'chunks', with lengths uniformly selected between 3 and 18 datapoints. For each chunk, there was a 10% probability of being flat and a 10% probability of inverting the parent region's slope sign. A chunk-specific slope multiplier from 0.01 to 1.99 is uniformly selected to further modulated the slope by multiplying the primary regions slope. Each chunk has a 50% probability of exhibiting linear drift (values by linear interpolation) or quadratic drift. For quadratic drift, a 'bend factor' from (0.75, 0.95) ∪ (1.05, 1.25) determined the midpoint deviation relative to an equivalent linear segment, and a quadratic function was fitted using the start, end, and modified midpoint. This restriction to linear and quadratic forms, combined with small, variable chunk sizes, generates diverse local baseline morphologies, mitigating overfitting, while subsequent noise addition ensures no segment remains perfectly ideal. This process ensures representation of baseline drift from mild to extreme (**Figure S1**), including features like plateau transitions and minor local slope reversals.

Two sets of global slope coefficients were used: [0.1, 1, 3, 6, 9, 14, 20, 27, 35] for denoising and baseline removal networks, and [0.2, 1, 3, 5, 7, 9] for peak parameter prediction networks. For context, a global slope coefficient of 35 yields a theoretical maximum drift-to-peak height ratio of approximately 2.96, while a coefficient of 9 results in a ratio of approximately 0.76, assuming ideal multipliers and no flat regions or inversions; actual average ratios are lower due to probabilistic elements. Including baseline drift for peak parameter models enhances robustness, as experimental baseline correction is seldom perfect, and probabilistic elements ensure some spectra exhibit flat baselines even with non-zero global coefficients. Finally, following the generation of peaks and baseline, Gaussian noise was added, with its standard deviation randomly selected from [0, 0.0001, 0.0025, 0.005] as a fraction of the noise-free maximum peak intensity.

**Data augmentation on baseline drift.** To further improve OASIS's comprehension of global spectral shapes and its ability to handle abrupt baseline discontinuities, a 'baseline shift' augmentation was incorporated. This augmentation has a 12% probability of occurring once within a spectrum and a 4% probability of occurring twice. Each shift transpires over a randomly determined number of data points, ranging from 2 to 5. A key characteristic of these shifts is that 65% to 90% of the total shift magnitude occurs between a single pair of adjacent data points within the shift region, with the remaining change distributed randomly among the other points in that region. Furthermore, a 'bounce' artifact can occur immediately before and after a shift, manifesting as a +/- 10% modification of the point's intensity relative to its current value. The magnitude of the overall baseline shift itself is uniformly randomly selected from the interval [0.05, 1.0]. This value is applied to the spectra after normalization, resulting in an effective shift of 2.5% to 50% of the total spectral amplitude range.

The synthetic baseline drift, as illustrated in **Figure S2**, was designed to emulate a wide range of experimentally observed phenomena, including those characteristics of UV-Vis absorption, fluorescence, and Raman spectroscopy. The generated baselines can exhibit rapid changes at spectral edges or within central regions while remaining relatively stable elsewhere; conversely, they can also display consistent trends across the entire spectrum or demonstrate broad reversals. Locally, the drift can flatten or reverse over short intervals, mimicking the nuanced behavior of real baseline drift **Figure S2a** demonstrates how the general trend of the drift reverses direction around the midpoint, while local trends can also reverse within these broader patterns. To enhance model robustness, various noise augmentations were systematically applied: **Figure S2b** incorporates beta 1 noise to train OASIS in recognizing that baseline drift need not conform to mathematically smooth functions, while **Figure S2c** applies beta 2 noise with larger spikes to further strengthen the neural network's sensitivity to experimental spectral variations. **Figure S2d** introduces baseline shifts, training OASIS to handle step-like increases or decreases in baseline levels. **Figure S2e** presents a comprehensive example featuring a new baseline drift with all previously described augmentations—beta 1 noise, beta 2 noise, and baseline shifts—additionally incorporating flat regions to represent another common experimental scenario. These noise augmentations are described in greater detail in the data augmentation on noise simulation section. Ultimately, OASIS becomes robust beyond the synthetic data to experimental data by being trained on such a broad range cases and augmentations. The Voigt profile-based spectra and the baseline drift components were generated independently, then combined and normalized prior to their use in model training. Two distinct parameter sets were employed for training different neural network modules: one for the denoiser and baseline removal network, and another for the peak location and intensity/FWHM selection network, with key differences pertaining to the range of slope coefficients for baseline drift and the standard deviations of added noise.

A quantitative comparison of baseline drift magnitudes can be made by assessing the ratio of the average maximum drift to the maximum peak height. For instance, a global slope coefficient of 35 yields a theoretical maximum drift-to-peak height ratio of 2.961, while a coefficient of 9 results in a ratio of 0.7614; these calculations assume all subgroup slope coefficients average to 1 and that the drift sign remains consistent without flattening throughout a section, an idealized scenario with a low (<4%) probability of occurrence across an entire section, thus the actual average drift-to-height ratios in the dataset are considerably lower. Specifically, the denoiser and baseline removal models were trained using global slope coefficients of [0.1, 1, 3, 6, 9, 14, 20, 27, 35], whereas the peak location and intensity/FWHM models utilized coefficients of [0.2, 1, 3, 5, 7, 9]. The deliberate inclusion of baseline drift in these spectral models aimed to enhance training robustness, as residual baseline artifacts are common in experimental data due to imperfect correction.

**Data augmentation on noise simulation.** To enhance model robustness and generalization, synthetic spectra were augmented with both Gaussian noise and noise derived from a beta distribution, the latter generated using NumPy's **random.beta** function with parameters alpha = 1 and beta = 2 (**Figure S1**). For Gaussian noise in the dataset to train the denoising model, standard deviations of [0, 0.001, 0.005, 0.01, 0.02, 0.05, 0.1, 0.15] relative to the maximum peak intensity were utilized. For models dedicated to baseline removal, peak finding, and peak fitting, a distinct set of Gaussian noise standard deviations, [0, 0.0001, 0.0005, 0.001], was applied. Regarding beta distribution noise, the denoiser and baseline remover shared the same application protocol: two independent sets of beta noise were applied, each with a 50% probability of inclusion per spectrum. The first set introduced noise with magnitudes ranging from 0.001 to 0.004, while the second set spanned magnitudes from 0.005 to 0.025. The beta distribution (alpha=1, beta=2) was chosen as it provides good representation across all noise values within these ranges, with larger magnitudes being inherently less probable. For the peak fitting and peak detection models, only the smaller magnitude beta noise set (0.001 to 0.004) was included; the larger magnitude set (0.005 to 0.025) was omitted.

The incorporation of noise, particularly non-Gaussian noise, beyond the denoising model was a deliberate strategy to prevent overfitting and improve the generalization capabilities of OASIS. This approach, including the differential application of beta noise to the final two model types (peak detector and peak fitter), was informed by several observations. Qualitative analysis of experimental spectra, especially in low-noise environments, revealed spectral artifacts, descriptively termed 'dents', which do not conform to simple Gaussian or Poisson distributions. These artifacts typically affect less than 20% of spectral points and are generally mild, covering only a few data indices, justifying the inclusion of the first (smaller magnitude) beta distribution. The second, larger magnitude beta distribution was introduced to simulate more extreme instances of such artifacts, thereby training OASIS to be more robust against noise profiles beyond purely Gaussian characteristics, to which networks can easily overfit. For instance, in synthetic data mimicking Raman spectra, if only substantial Gaussian noise is added, the network may proficiently identify low-intensity peaks but can also exhibit a tendency to erroneously identify (or 'hallucinate') small, sharp peaks in noisy experimental data. The addition of the second beta distribution mitigates this by inducing a more conservative peak prediction behavior in challenging noise environments, encouraging the network to refrain from predicting a peak if uncertainty is high.

The second beta distribution was reintroduced to further enhance robustness of baseline correction model; this forces the network to discern the 'general shape' of a spectrum to accurately locate the baseline, improving its adaptability to diverse and unusual experimental conditions. The first beta distribution was retained for the baseline remover as the denoising model may not perfectly eliminate all 'dent'-like artifacts. For the peak detector and peak fitter, Gaussian noise and the first beta distribution were maintained for robustness. However, the second beta distribution was excluded because these models are intended for refined and precise parameter extraction rather than learning broad spectral shapes. Overall, this multi-faceted noise augmentation strategy was designed to encourage the initial sequence of models (denoiser, baseline remover, peak detector) to disregard peaks that are excessively obscured by noise or artifacts..

**Data augmentation on peak location and FWHM.** To further enhance the robustness of the peak selection and intensity/FWHM models and mitigate potential overfitting to idealized Voigt profiles, a 'spectral tweaking' augmentation was implemented. While these models are not subjected to the larger magnitude beta noise to preserve precision, additional generalization beyond Gaussian noise and the first (smaller magnitude) beta noise was deemed beneficial. Spectral tweaking involves the targeted modification of intensity values at the identified peak indices and at a limited number of adjacent points. This augmentation was applied to the training data for both the peak fitter and peak detector models. The probabilistic application framework consists of an initial 50% probability for spectral tweaking to be active for a given spectrum. Subsequently, for each peak within an activated spectrum, an independent 50% probability determines if tweaking occurs for that specific peak. If selected, the intensity at the peak's central index is multiplied by a factor uniformly sampled from the union of intervals (0.96, 0.999) U (1.001, 1.04). For points flanking the peak, a random integer, uniformly selected from [0, 3], determines the number of adjacent points on each side of the peak that are modified. The intensities of these selected adjacent points are multiplied by a factor uniformly sampled from (0.98, 0.999) U (1.001, 1.02). Peak shadowing

A 'peak shadowing' augmentation was introduced specifically for the peak fitter model to improve its robustness to minor inaccuracies in input peak locations. This augmentation has a 50% probability of being enabled for an entire spectrum, and if enabled, each individual peak within that spectrum has a further 50% probability of undergoing shadowing. For peaks selected for this augmentation, the labeled datapoint fed to the model as the peak location is shifted by 1, 2, or 3 data points, with the magnitude and direction (left or right) of the shift being uniformly and randomly selected. Crucially, this augmentation modifies only the input label provided to the peak fitter model; the underlying spectral data remains unchanged. Peak shadowing enables the model to learn to predict the maximal intensity and FWHM of a peak even if the input peak location (as determined by the preceding peak detector) is slightly offset from the true peak maximum. This allows the intensity and FWHM estimation model a degree of flexibility to refine these parameters based on the local spectral morphology around a slightly misaligned input coordinate, rather than being rigidly constrained by it. The peak fitter model is not trained to re-identify peak locations or modify input spectral data; this augmentation is solely designed to provide minor operational tolerance in fitting peak contributions.

*Model architecture and training*

**Early Architecture Exploration**. Initial conceptualization explored a two-network system: one for concurrent denoising and baseline removal, and a second for simultaneous peak location, intensity, and Full Width at Half Maximum (FWHM) determination. However, extensive experimentation with numerous model architectures revealed superior performance when each task was addressed by a dedicated, specialized model. This modular approach also facilitated more effective fine-tuning of individual components, as optimal hyperparameters (e.g., kernel/filter sizes, activation functions) were found to be task-specific. For instance, distinct kernel/filter size optima were identified for denoising versus baseline removal, and similarly for peak location versus intensity/FWHM prediction. These observations held true for both simpler Convolutional Neural Network (CNN) designs and the more complex U-Net architecture. An alternative looped system, involving six distinct neural network architectures, was also prototyped. This system proposed an initial peak location and intensity prediction, followed by baseline removal informed by these predictions, and concluding with a refined peak location and intensity prediction. However, validation losses for this more complex system did not surpass those of the simpler, four-stage sequential approach during initial prototyping, and thus this line of inquiry was not pursued further, though it may represent a promising avenue for future research. Early attempts at combined-task models further confirmed that dedicated architectures for each specific output type (e.g., denoising, baseline removal, peak parameters) performed significantly better due to task-specific optimal filter/kernel combinations.

Architectural explorations for peak location were initially inspired by Li et al. [21], featuring a simple CNN with kernel sizes k = [[11,1,11,1,1,11,1,3]] and filter counts f = [[40,20,10,20,10,30,18,18]], ReLU activation, and a single max-pooling operation. Inputs included the raw spectrum and its derivatives, with outputs for peak location (classification with cross-entropy) and regression (MSE). Adding dense layers was investigated but did not outperform this simple CNN, and various other filter/kernel combinations also performed worse. The Vicinity Peak Response (ViPeR) loss function was developed to encourage deeper pattern learning beyond what BCE + MSE allowed. A U-Net style architecture was then explored, finding that removing sampling and skip operations around 1x1 kernels improved performance. After testing GELU, Mish, Swish, and ReLU, GELU was selected as the best performing activation function for peak location.

For the Intensity and FWHM architecture, the initial design mirrored the early peak picker architecture, with two outputs and peak location data as an additional input. MSE performed better than MAE as a loss function, and weighted loss functions did not improve validation MSE during these initial trials.

Development of the Baseline Removal model began with a simple CNN architecture. An early promising configuration used LeakyReLU activation, kernel sizes k = [[3, 5, 7, 9, 1, 11]], and filters f = [[18, 20, 22, 24, 20, 26]], with MSE as the loss function. Incorporating dense layers proved detrimental. A U-Net style architecture with up/downsampling and skip connections improved validation metrics. Further refinement showed that removing skip connections and up/downsampling around 1x1 kernels, similar to the peak picker, further enhanced performance. Attempting to use the peak detection kernel/filter combination for baseline removal was unsuccessful, highlighting the distinct requirements of these tasks. After testing LeakyReLU, GELU, Swish, Mish, and ISRLU, Mish was found to be optimal for baseline removal with a validation MAE of 0.0144. Renewed attempts to integrate dense blocks were unsuccessful, and customized loss functions with attention-like mechanisms did not yield significant improvements. Smaller kernel sizes generally performed worse during these explorations.

For the denoising model, various filter and kernel combinations were tested within a U-Net architecture. MSE was found to be superior to MAE as a loss function, and Swish was determined to be the best performing activation function among those tested (ReLU, LeakyReLU, GELU, Swish, Mish, ISRLU).

**Final Model Design** Based on early experimentation, the final unified approach comprises four neural networks operating in series: (1) denoising, (2) baseline removal, (3) peak location, and (4) combined intensity and FWHM determination. As depicted in Figure S1, U-Net architectures incorporating residual learning and multi-stream processing were developed for all four tasks, sharing several common design elements. Each network accepts a minimum of three input channels: the raw spectrum, its first derivative, and its second derivative. Each input stream is processed through an independent U-Net pathway composed of residual blocks. Each residual block consists of two 1D convolutional (Conv1D) layers with batch normalization, utilizing variable activation functions and dropout layers (with a drop-path mechanism during training). The processed streams from the three U-Net pathways are subsequently merged via concatenation, followed by a max-pooling operation with a stride of 2. The U-Net structure consistently outperformed simpler CNNs for noise and baseline removal, and for peak location and intensity/FWHM determination, a hybrid U-Net structure yielded the best results. L1 and L2 regularization were employed on all final trained models.

The Denoising Network achieved optimal performance with kernel sizes of k = [[3, 5, 7, 9]] and filter counts of f = [[17, 19, 21, 23]], using Swish as the activation function. This configuration allows the network to consider local spectral features for noise identification, with the output being the denoised spectrum.

The Baseline Removal Network utilized kernel sizes k = [[5, 7, 9, 11, 11, 15]] and filter counts f = [[20, 22, 24, 26, 22, 28]], with Mish as the activation function. This enables capture of features across multiple scales to identify the general baseline trend; its output is the predicted underlying baseline drift.

The Peak Location Network employed a hybrid U-Net architecture where skip connections are omitted around 1x1 kernels, used GELU as the activation function, and had optimal kernel sizes k = [[15, 1, 15, 1, 1, 15, 1, 4]] and filter counts f = [[20, 20, 6, 20, 10, 17, 18, 14]]. This design was partially inspired by Li et al. [21] for its structure, facilitating broad context and detailed local analysis.

The Intensity and FWHM Network also used a similar hybrid U-Net architecture, adopting Swish as the activation function, with optimal kernel sizes k = [[15, 1, 15, 1, 1, 15, 4]] and filter counts f = [[24, 20, 8, 20, 10, 20, 18, 16]]. This network takes the original spectrum and predicted peak locations as inputs.

Following peak location prediction, a classical non-maximum suppression method is applied as a post-processing step. Predicted peak locations (values between 0 and 1) are thresholded at 0.5. Within a sliding window of 6 data indices, only the peak with the highest prediction confidence exceeding the 0.5 threshold is retained, while others are suppressed, refining the output for the subsequent intensity and FWHM determination model.

The observation that each of the four chosen models excels at its designated task while performing sub-optimally on others strongly suggests potential for further performance enhancements through more integrated in-series, shared, or in-parallel architectural designs, though practical implementation presents considerable challenges.

**Input data variation and model performance.** The decision to utilize a multi-channel input, comprising the raw spectrum, its first derivative, and its second derivative, for OASIS, was based on empirical evidence demonstrating improved performance. As presented in **Figure S3**, an investigation was conducted to assess the impact of different input combinations on model efficacy. These evaluations, performed early in model development using the peak locator neural network trained on a synthetic dataset specifically devoid of noise or baseline drift, provided clear insights. The results showed that the lowest final validation loss (40.3) was achieved when the network was provided with all three input channels concurrently: the raw spectrum, its first derivative, and its second derivative (**Figure S3a**). In contrast, providing only the raw spectra yielded a higher validation loss of 48.2 (**Figure S3b**). Augmenting the raw spectra with only the first derivative resulted in a validation loss of 48.7 (**Figure S3c**), while combining it with only the second derivative led to a validation loss of 44.6 (**Figure S3d**).

The superior performance observed even on this idealized, noise-free dataset strongly suggests that providing these mathematically derived features explicitly aids the network in more robustly identifying and localizing peaks. The first derivative provides explicit information regarding the local slope of the spectrum, effectively highlighting inflection points and enabling the network to better discern peak edges, shoulders, and resolve closely spaced features. The second derivative accentuates peak curvature, which can effectively sharpen the appearance of peaks and make their precise locations more distinct, particularly for broader peaks. By supplying the raw spectrum alongside these derivatives, the network is furnished with a richer, more diverse set of features representing the intensity, rate of change, and curvature of the spectral signal at each point. This multi-faceted input likely enables the network to learn a more comprehensive and discriminative mapping from input to output. It potentially reduces the learning burden on the network, as it does not need to implicitly derive these fundamental shape characteristics solely from the raw data, which can be a more complex task. The distinct information content of each channel likely contributes to a more robust feature extraction process within the initial layers of the neural network. The observation that this combination is beneficial even in the absence of noise or complex baselines indicates a fundamental advantage in how the network processes and interprets spectral information. It is anticipated that this enhanced feature representation will confer even greater benefits when the models are applied to more challenging experimental data, where noise and baseline variations can obscure subtle peak characteristics that the derivatives might help to disambiguate. Consequently, this multi-channel input strategy (raw spectrum, first derivative, and second derivative) was adopted as a standard for all subsequent model development and training due to its demonstrated ability to improve model performance and likely contribute to overall model robustness and generalization.

**Custom loss functions for denoising.** The loss function employed for the noise remover network was a dynamically weighted Mean Squared Error (MSE). The training data for this model included both Gaussian noise and, potentially, shot noise components that required removal. The dynamic weighting of the MSE loss was specifically determined by the standard deviation of the added Gaussian noise. The possible Gaussian noise standard deviations (std) used in training were [0.001, 0.01, 0.02, 0.05, 0.1, 0.15].

The corresponding multiplicative coefficients applied to the MSE loss for these respective noise levels were [100, 40, 15, 4, 1, 1].

As illustrated in **Figure S4**, this dynamically weighted loss function demonstrated superior performance on spectra with low noise levels compared to a standard MSE where the weighted factors are all 1.0. **Figure S4a** shows the denoising results using the dynamic weighting approach on the full spectrum, while **Figure S4b** provides a zoomed-in view highlighting the performance on a specific spectral region, and **Figure S4c** presents a detailed comparison between ground truth and denoised spectra. In contrast, **Figure S4d** demonstrates the results obtained using standard MSE loss with equal weighting on the full spectrum, **Figure S4e** shows the zoomed-in view of the equally weighted approach, and **Figure S4f** presents the detailed comparison for the standard MSE method. The zoomed-in portion of the spectra in Figure S4e clearly shows that an equally weighted loss function ceased to improve earlier in training, whereas the dynamic weighting allowed for continued optimization, particularly on low-noise spectra as shown in **Figure S4b** and **S4c**. This extended training, driven by improvements on less noisy data, effectively prevents over-smoothing. The prevention of over-smoothing is critical due to the sequential operation of OASIS's pipeline; the subsequent baseline remover module requires input data that closely resembles the synthetic training data distribution, a condition facilitated by the dynamic weighting strategy as seen in **Figure S4a, S4b,** and **S4c** compared to the standard approach shown in Figure **S4d, S4e,** and **S4f.Custom loss function for baseline correction.** The loss function for the baseline remover network was a composite function designed to primarily minimize the MSE of the predicted baseline while concurrently promoting smoothness in the output, thereby preserving the local features of the spectra after baseline subtraction (**Figure S5**). While employing solely MSE loss can yield a marginally lower MSE value for the baseline itself compared to the optimized composite function, the latter provides Figure S5a demonstrates the baseline correction performance using optimal smoothing coefficients, displaying the raw spectra alongside the ground truth baseline and the neural network's predicted baseline. The optimal coefficients were determined through a combined optimization approach that minimizes both mean squared error (MSE) and signal recovery artifacts, as detailed below. Figure S5b illustrates the baseline prediction results when high smoothing coefficients are applied, showing improved shape preservation in certain spectral regions but with substantial MSE deviations in others. In contrast, Figure S5c presents the baseline correction output from a network trained exclusively with MSE loss (unsmoothed approach), which exhibits lower computational smoothing but compromises signal quality. Figure S5d compares the recovered signals obtained through different approaches: the ground truth signal, the signal recovered using ideal smoothing parameters, and the signal recovered without smoothing regularization. While the unsmoothed approach yields marginally lower MSE values, the recovered signal exhibits excessive noise and fails to preserve the characteristic spectral features as effectively as the ideally smoothed neural network output. The high smoothing configuration demonstrates superior shape preservation in select cases; however, the significant MSE increases across multiple spectral regions led to the selection of balanced parameters that optimize both MSE minimization and peak shape fidelity. two significant advantages. Firstly, when applied to experimental data where the true baseline may not be perfectly flat beneath peaks, the model trained with the composite loss function produces a smoother predicted baseline in such instances, and peak intensities are consequently better preserved. In contrast, models trained solely with standard MSE loss exhibit more pronounced deviations in these challenging scenarios. Secondly, the local topography of peaks post-subtraction is notably smoother when using the composite loss function.

**Custom loss function for peak selection: ViPeR**

To improve training efficiency and accuracy for spectral peak identification, especially with limited datasets, class imbalance and sharp loss environments, we introduce ViPeR loss.

$$V = F_P + P_V + P_L \tag{1}$$

$$F_p = \alpha \sum_i \begin{cases} Y_P^{(i)} \cdot (d^{(i)} - v)^2 & \text{if } d^{(i)} > v \\ 0 & \text{if } d^{(i)} \leq v \end{cases} \tag{2}$$

$$P_V = \begin{cases} \beta \sum_i (Y_P^{(i)} - 1)^2 & \text{if } Y_T^{(i)} = 1 \\ \beta \sum_i (Y_P^{(i)} - 0)^2 & \text{if } Y_T^{(i)} = 0 \end{cases} \tag{3}$$

$$P_L = \gamma \sum_{i=1}^{n} \begin{cases} (Y_P^{(i)} - 1)^2 \cdot \left(1 - \frac{d^{(i)}}{v}\right)^2 & \text{if } d^{(i)} \leq v \\ 0 & \text{if } d^{(i)} > v \end{cases} \tag{4}$$

A key advantage of ViPeR is its ability to provide smooth gradient signals near spectral peaks, enabling neural networks to first learn general peak patterns and then refine localization accuracy during training. The total ViPeR loss, V, is calculated as the sum of three distinct components: the False Positive Penalty ($F_P$), the Vertical Penalty ($P_V$), and the Lower Penalty ($P_L$). For each predicted point i, the model output is $Y_P^{(i)}$, and $d^{(i)}$ is the distance to the nearest ground-truth peak center. The parameter $v$, a dynamic vicinity threshold that defines a spatial window around true peaks; this threshold $v$ defines a dynamic vicinity threshold around true peaks and is annealed during training. Each loss component is weighted by coefficients $\alpha$ for $F_P$, $\beta$ for $P_V$, and $\gamma$ for $P_L$.

The $F_P$ component (**Eq. 2**) becomes nonzero only for predictions where $d^{(i)}$ exceeds $v$. In such cases, the penalty for a given point *i* is the product of its predicted intensity $Y_P^{(i)}$ and the squared excess distance, $(d^{(i)}-v)^2$. The $P_V$ component (**Eq. 3**) compares the predicted intensity $Y_P^{(i)}$ to the ground truth label $Y_T^{(i)}$, which is defined as 1 if the point *i* is a true peak center and 0 otherwise. $P_V$ accumulates the squared differences $(Y_P^{(i)} - Y_T^{(i)})^2$ over all points. Finally, the $P_L$ component (**Eq. 4**) is active for predictions within the vicinity of a true peak, that is, where $d^{(i)} \leq v$. For these points, the penalty is calculated as the product of the squared difference $(Y_P^{(i)}-1)^2$ and a modulating factor $(1-d^{(i)}/v)^2$. This factor scales the penalty based on the prediction's normalized distance from the true peak center within the vicinity v. The sum of these three weighted components constitutes the final loss *V*, guiding the model's optimization.

The modular architecture of the ViPeR loss affords significant flexibility in tailoring the optimization objective. By adjusting the weighting coefficients ($α$, $β$, $γ$), researchers can strategically prioritize distinct aspects of model performance—such as the stringent suppression of false positives (via $α$) or the refinement of sub-pixel localization accuracy (via $γ$)—a level of control not available in monolithic loss functions. Furthermore, the inclusion of the dynamic vicinity threshold, $v$, makes the loss function particularly amenable to curriculum learning strategies. This structure allows for a coarse-to-fine training regimen, implemented through a dynamic or custom annealing schedule, where the model first learns to identify general peak regions before honing localization precision. This adaptability enables the training process to be precisely matched to the unique statistical challenges of a given dataset, ultimately fostering more robust convergence and superior model generalization.

**Custom loss function for peak intensity and FWHM.** The loss function for the peak intensity prediction model was custom-designed to encourage accurate and complete prediction of well-defined peaks (**Figure S6** and **S5**). Standard MSE loss, due to its mathematical nature and the characteristics of the spectra simulation (where baseline drift contributes positively to apparent peak height), tends to incentivize OASIS to under-predict peak intensities as a conservative strategy. The custom loss function was developed to counteract this tendency. While this custom loss function results in a negligible increase in the MSE of the predicted intensity itself, it yields approximately a 25% improvement in the MSE for FWHM prediction. More importantly, when the neural networks comprising OASIS are used in series, this custom loss function enables the model to predict peak intensities with greater accuracy. An illustrative, albeit uncommon, example is highlighted in **Figure S6a to S6c** and **Figure S7a and S7b**, where a network trained on standard MSE loss exhibits confusion when applied to synthetic data that has undergone prior noise and baseline correction by upstream models. As shown, the network trained with the custom loss function achieves minimal error in all presented cases. In contrast, the MSE-trained network significantly misestimates one of the peaks and fails to predict the FWHM for a small peak at 383 nm misses one of the peaks by a large degree and fails to even attempt to predict the FWMH of a small peak at 383.

**Human vs. OASIS Comparison**

OASIS results were compared to human expert predictions on three sets of experimental data using 0.5 M 9-fluorenone-2,7-dicarboxylic acid in 3M KOH (Raman), 0.04 M $TbCl_3$ aqueous solution (fluorescence), and 0.08 M $Nd_2(SO_4)_3$ aqueous solution (UV-Vis). Human expert annotation for intensity and FWHM was not performed for highly convoluted peaks (e.g., at 1698.0 cm$^{-1}$ and 1711.3 cm$^{-1}$ in Table S4, where "Convoluted Peak" is noted for expert values), however they comport well by visual inspection. For Raman data (**Table S4**), the model's predictions for well-resolved peaks generally showed good agreement. For instance, at 1605 cm$^{-1}$, the model predicted an intensity of 137.6 A.U. and FWHM of 9.6 cm$^{-1}$, closely matching the human annotation of 156.0 A.U. and 9.5 cm$^{-1}$ respectively (normalized error of 0.12). However, for small peaks, the model consistently predicted lower intensities; for example, at 648.3 cm$^{-1}$, the model's intensity was 1.2 A.U. compared to the expert's 2.7 A.U. FWHM values for these small peaks also diverged, with the model predicting 1.8 cm$^{-1}$ FWHM at 648.3 cm$^{-1}$ versus the expert's 4.5 cm$^{-1}$. However, the model still identifies these peaks in a consistent manner which indicates application to automatically analyze spectra is highly promising, and the overall normalized error is generally lower than 0.1 for intensity. Fluorescence (**Table S5**) and UV-Vis (**Table S6**) data were also analyzed. The most notable difference was in the fluorescence data (**Table S5**) at 553.2 nm, where OASIS predicted an intensity of 0.06 A.U. and a substantially wider FWHM of 18.36 nm, compared to the expert's annotation of 0.02 A.U. intensity and 1.50 nm FWHM for this very small, partially overlapping peak. This peak is highly ambiguous and miniscule, which causes a lot of ambiguity no matter how the analysis is done. For the better resolved peaks OASIS and expert results are in close agreement on intensities. The fluorescence analysis highlights the pipeline's robustness; while the baseline correction model did not fully flatten a broad baseline feature between 600–650 nm in the fluorescence spectra, the downstream peak identification model correctly assigned no peak to this region, preventing a cascading error. The human expert determined there was no peak in this region by considering subsequent time resolved spectra. Regarding denoising, OASIS removed an artifact similar to a large spike induced artifact in the fluorescence data as seen in **Figure 5** of the main manuscript despite this artifact having no equivalent in the training augmentation methods. OASIS effectively corrected significant and abrupt baseline jumps in **Figure 5** which is a common and difficult challenge for automated processing, however the synthetic data had augmentations for this type of experimental error. Further demonstrating robustness, OASIS successfully was applied to experimental Raman spectra with 18 peaks despite being trained on synthetic data with a maximum of 11 peaks.

**Data availability**

Data is available upon reasonable request.

**Code availability**

The code used in this study is not publicly available due to an application of intellectual property (IP) protection currently under review. The authors are in the process of developing the code into a product that will be made available for future use. Although the source code cannot be shared at this time, the authors are happy to assist others with data analysis upon reasonable requests.

**Acknowledgements**

This material is based on work supported by the Advanced Research Projects Agency-Energy's (ARPA-E) Mining Innovations for Negative Emissions Resource Recovery (MINER) program with award number 22CJ0000901, "Re-Mining Red Mud Waste for $CO_2$ Capture and Storage and Critical Element Recovery (RMCCS‐CER)". KMR and XZ also acknowledge support from the U.S. Department of Energy (DOE), Office of Science, Basic Energy Sciences (BES), Chemical Sciences, Geosciences, and Biosciences Division through its Geosciences Program at Pacific Northwest National Laboratory (PNNL) (FWP 56674). The Graduate Fellow, C.Y., was supported by the DOE, Office of Environmental Management—Minority Serving Institutions Partnership Program (EM MSIPP). Part of the research was performed with a user proposal #61223 (Award DOI: 10.46936/lsr.proj.2024.61223/60012698) and 51922 (Award DOI: 10.46936/lser.proj.2021.51922/60000373) at the William R. Wiley Environmental Molecular Sciences Laboratory (EMSL), a national scientific user facility sponsored by the U.S. DOE's Office of Biological and Environmental Research and located at PNNL in Richland, WA. PNNL is a multiprogram national laboratory operated by Battelle Memorial Institute under contract no. DE-AC05-76RL01830 for the DOE.

**Authors' contributions**

C.Y., J.L., and X.Z. conceived and designed the project. C.Y., J.L., and E.L. developed the code, implemented the model, and performed data analysis. Y.F. and M.L.M. carried out the spectroscopy experiments under the supervision of Z.W. C.Y., J.L., and X.Z. drafted the manuscript. X.Z. supervised the overall project. All authors contributed to the discussions and revisions of the manuscript.

**Competing Interests**

The authors declare no competing interests.

**Additional information**

Supplementary information: The online version contains supplementary material available at https://doi.org/xxxxx.

Correspondence and materials requests should be addressed to Xin Zhang (xin.zhang@pnnl.gov).

**Peer Review Information**

Nature Machine Intelligence thanks xx, and the other, anonymous, reviewer(s) for their contribution to the peer review of this work.

Reprints and permission information is available at www.nature.com/reprints.